\documentclass[letterpaper, 10 pt, conference]{ieeeconf}  

\IEEEoverridecommandlockouts                              

\usepackage{comment}
\let\labelindent\relax
\usepackage{enumerate}
\usepackage{gensymb}
\usepackage{algpseudocode}
\usepackage{algorithm}
\usepackage{xcolor}
\usepackage{multirow}
\usepackage{subcaption}
\usepackage{multirow}
\usepackage{graphicx}
\usepackage{diagbox}

\newcommand{\figref}[1]{Fig.~\ref{#1}}

\newcommand{\tabref}[1]{Tab.~\ref{#1}}

\usepackage{enumitem}

\newcommand\B[1]{\textcolor{blue}{#1}}

\makeatletter
\let\NAT@parse\undefined
\makeatother

\usepackage[pagebackref=true,breaklinks=true,colorlinks,bookmarks=false,citecolor=green]{hyperref}

\title{\LARGE \bf
LiDAR-CS Dataset: LiDAR Point Cloud Dataset with Cross-Sensors \\for 3D Object Detection
}

\author{Jin Fang$^{1,2}$, Dingfu Zhou$^{2}$, Jingjing Zhao$^{2}$, Chenming Wu$^{2}$, \\ Chulin Tang$^{3}$, Cheng-Zhong Xu$^{1}$, and Liangjun Zhang$^{2}$
\thanks{ $^{1}$ State Key Lab of IOTSC, CIS, University of Macau.}
\thanks{ $^{2}$ Robotics and Autonomous Driving Laboratory, Baidu Research.}
\thanks{ $^{3}$ University of California, Irvine.}
}

\begin{document}

\maketitle
\thispagestyle{empty}
\pagestyle{empty}

\begin{abstract}

Over the past few years, there has been remarkable progress in research on 3D point clouds and their use in autonomous driving scenarios has become widespread. 
However, deep learning methods heavily rely on annotated data and often face domain generalization issues. 
Unlike 2D images whose domains usually pertain to the texture information present in them, the features derived from a 3D point cloud are affected by the distribution of the points. 
The lack of a 3D domain adaptation benchmark leads to the common practice of training a model on one benchmark (e.g. Waymo) and then assessing it on another dataset (e.g. KITTI). This setting results in two distinct domain gaps: scenarios and sensors, making it difficult to analyze and evaluate the method accurately.
To tackle this problem, this paper presents \textcolor{red}{\textbf{LiDAR}} Dataset with \textcolor{red}{\textbf{C}}ross-\textcolor{red}{\textbf{S}}ensors (LiDAR-CS Dataset), which contains large-scale annotated LiDAR point cloud under six groups of different sensors but with the same corresponding scenarios, captured from hybrid realistic LiDAR simulator. To our knowledge, LiDAR-CS Dataset is the first dataset that addresses the sensor-related gaps in the domain of 3D object detection in real traffic. Furthermore, we evaluate and analyze the performance using various baseline detectors and demonstrated its potential applications.
Project page: \href{https://opendriving.github.io/lidar-cs}{\textit{https://opendriving.github.io/lidar-cs}}.
\end{abstract}


\section{Introduction}\label{sec:in}

In recent years, deep learning techniques have advanced quickly, and the hardware that supports them has also improved. As a result, deep-learning-related applications have become more prominent, and one of the most noteworthy is Autonomous Driving (AD). To achieve 3D perception tasks, Light Detection and Ranging (LiDAR) sensors have become the leading choice due to their reliable range measurement.
To well promote the technique development, many AD-related 3D perception datasets have been released such as the KITTI \cite{Geiger2012CVPR}, nuScene \cite{caesar2020nuscenes} and WOD \cite{sun2020scalability}.

Despite the notable accomplishments, pressing concerns still confront the community. Evaluating models across different datasets can be a complex task, which requires careful consideration of various factors. Usually, this kind of evaluation is necessary to verify the model's generalization ability, however, this process is very complex and time-consuming due to several reasons: \textbf{(1) different sensors.} The LiDAR points are collected from different sensors, e.g., KITTI (Velodyne 64 beams LiDAR), nuScene (32 beams LiDAR); \textbf{(2)  different labeling rules.} The labeling rule is different, in the KITTI dataset, only the objects in the front views have been annotated while the nuScene and WOD are labeled in 360\degree view; \textbf{(3) different categories.} KITTI and WOD share the same labeled categories while nuScene defines more specific categories. \textbf{(4) different evaluation metrics.} Last but not least, the evaluation metric for each dataset is also quite different which confuses the comparison over different datasets.

\begin{figure}[t!]
	\centering
	\includegraphics[width=0.46\textwidth]{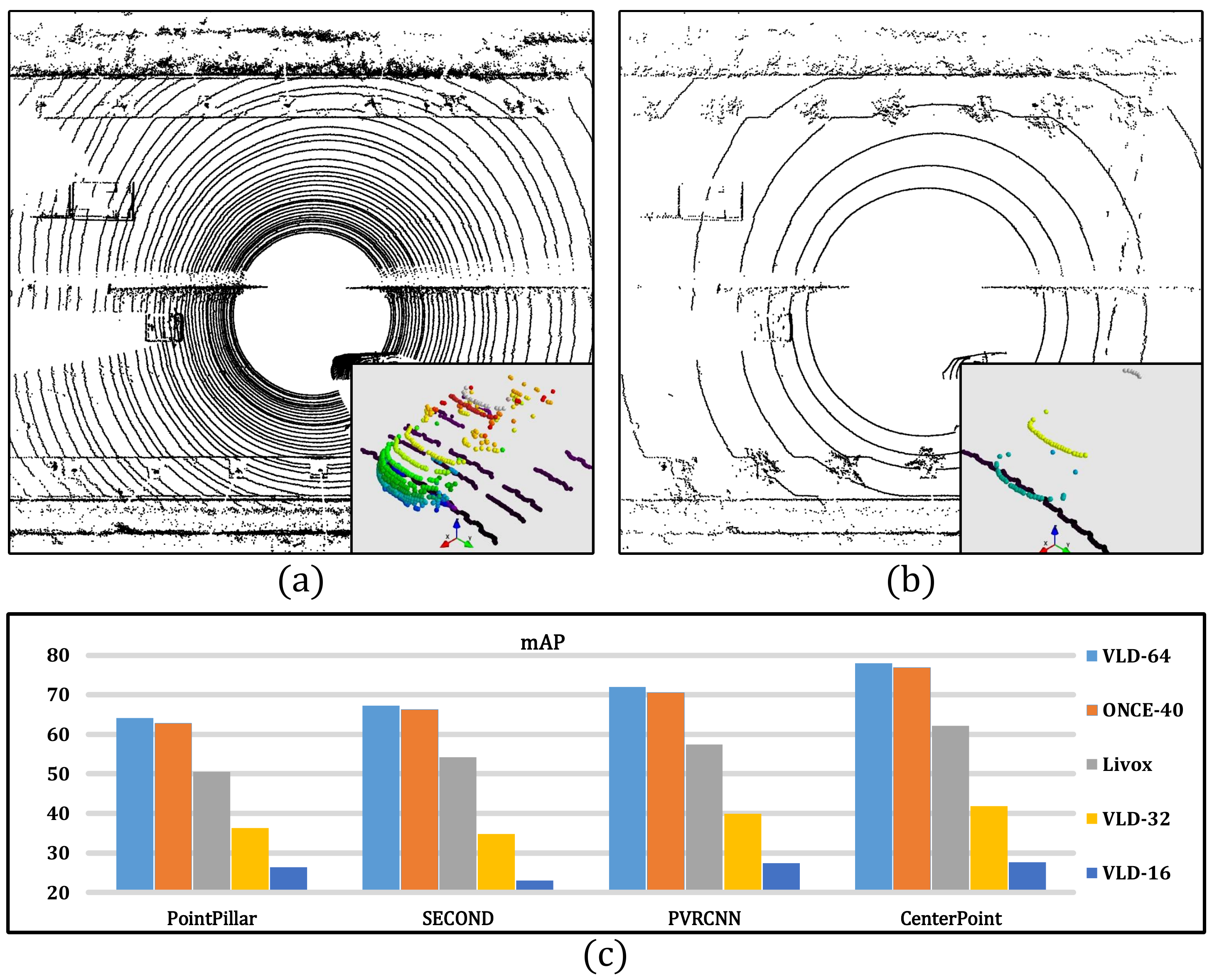} 
	\centering
	\caption{(a) and (b) are LiDAR point cloud examples collected from different types of sensors which are from 64-beams and 16-beams LiDAR sensors respectively. The vehicle has been cropped and zoomed in for detailed visualization purposes. Sub-figure (c) gives a cross-sensors evaluation of experimental results where four baseline detectors are trained on VLD 64 LiDAR data and evaluated on five different sensors in the same scenario. The results show that the domain gaps are obvious across different sensors.}
	\label{Fig:head_img}
\end{figure}

Secondly, there have been many different LiDAR sensor designs and functions due to the rapid development of autonomous driving technology recently, such as Velodyne \cite{link_velodyne}, Pandora \cite{link_hesai} and Livox \cite{link_livox}, For each manufacturer, there are lots of LiDAR sensor types. However, the existing LiDAR point cloud benchmarks are trying to enlarge the scenario diversity while the diversity of sensors has been ignored.  However, none of them provide more diverse data with different LiDAR sensors. As shown in \figref{Fig:head_img}, the detectors' performances drop dramatically when the training and the testing are based on different sensors. Therefore, a dataset that can enlarge the sensor's diversity will be very helpful for the community. 

Thirdly, domain adaptation is a crucial challenge in point cloud perception research. Previous work \cite{wang2020train,yang2021st3d,zhang2023uni3d} employed multiple datasets for the research of domain adaptation. However, one fact is often ignored both the scenario domain and sensor domain exist across different datasets. The existence of two variables at the same time makes it difficult to evaluate and analyze different domain adaptation approaches. In most cases, it is easy to collect data from various situations using a single sensor. However, it becomes challenging to obtain data with identical scenarios using different sensors in real-world applications. This is because the background may stay constant, while the foregrounds are constantly changing.

Motivated by the issues mentioned above, a novel LiDAR-CS (LiDAR-Cross-Sensors) dataset has been proposed here which can well address these issues. In general, compared with existing benchmarks, LiDAR-CS dataset has the following unique properties. 

\noindent\textbf{Cross Sensors.} Currently, LiDAR-CS includes samples from 6 types of LiDAR sensors. For each LiDAR frame, we have generated 6 duplicates with the same scenarios and the same foreground position, the only difference is the point distribution which comes from the inherent attributes of the sensors, such as the beam number, field of view, angle resolution, etc. In addition, the data can be easy to be extended by adding another LiDAR sensor while only needing to provide some real scanned frames.

\noindent\textbf{Consistent Annotation.} We generate all the sensors' data with the same background and foreground, therefore, we can keep consistent annotations for all the sensors. With the same annotation, the evaluation metric therefore can remain the same for all the data.

\section{Related Work} \label{sec:related_work}

\vspace{2mm}
\noindent\textbf{LiDAR Dataset for AD Perception.}
Recently, numerous large-scale datasets have been released due to the rapid development of deep learning and autonomous driving.
For the 3D object detection task, the most popular benchmarks are KITTI \cite{Geiger2012CVPR}, nuScenes \cite{caesar2020nuscenes}, and WOD \cite{sun2020scalability}. KITTI is the first large-scale public dataset for perception tasks in real AD scenarios. The point cloud is collected by a 64-beam Velodyne (VLD) LiDAR device. Three categories have been annotated in this benchmark while only the objects visible in the front camera view have been annotated. NuScenes is another dataset collected by a 32-beam LiDAR device. The point cloud frames are fully annotated with a 360$\degree$ view. Objects are divided into 23 classes with attributes like visibility. WOD \cite{sun2020scalability} was conducted using 1 mid-range LiDAR and 4 short-range LiDARs. The categories of objects follow KITTI, but the frames are annotated in the full 360\degree~view. 

\vspace{2mm}
\noindent\textbf{Cross Domains Dataset.}
Domain (or Sensor) generalizability is a very important property for deep learning-based approaches. A lot of benchmarks have been released for the 2D image domain \cite{Syn2real_2018} while few works have been proposed for the 3D point cloud. In \cite{rist2019cross}, a \textbf{cross-sensor} dataset has been proposed with two types of LiDAR sensors (e.g., HDL-64 and VLP-32) and the transferability of features extracted from different sensors are also discussed, but unfortunately, the dataset is not released to the public. WOD \cite{sun2020scalability} proposes a \textbf{cross-scenario} domain adaptation task, which spans data from three different cities and split the data into two different domains. Experiments show that remarkable domain gaps exist between those cities. 

\vspace{2mm}
\noindent\textbf{LiDAR Simulation.}
LiDAR simulation \cite{lee2015lidar} \cite{tallavajhula2018off} is very popular in robotics and AD. \cite{GTA_V_2018} collects the calibrated images and point cloud using the APIs provided by the video game engine, and applies these data for vehicle detection in their later work \cite{wu2018squeezeseg}. CARLA \cite{CARLA_2017carla} and AutonoVi-Sim \cite{best2018autonovi} also furnish the function to simulate LiDAR points data from the virtual world. Unlike previous simulators that entirely rely on CG models and game engines, in \cite{Augmented_Sim_2020}, an augmented simulator based on the ``scan-and-simulate'' pipeline has been proposed. Specifically, a vehicle with a LiDAR scanner is deployed to sweep the street to obtain the background points cloud and then the foreground objects (e.g., CAD models) are placed on the background. LiDARsim~\cite{manivasagam2020lidarsim} leverage the multi sweeps to build the dense background and reconstruct the foreground from cropped point cloud instead of CAD models. Nowadays popular NeRF \cite{mildenhall2021nerf} technique is introduced for LiDAR simulation \cite{li2023pcgen,tao2023lidar,zhang2023nerf}.

\vspace{2mm}
\noindent\textbf{LiDAR-based 3D Object Detection.}
LiDAR-based 3D object detection has been well-studied for many years. Generally, these approaches can be divided into three categories by the way of LiDAR point cloud representation as view-based~\cite{ku2018joint,luo2018fast,yang2018pixor,liang2020rangercnn,zhou2019iou,hu2020you}, voxel-based~\cite{zhou2018voxelnet,yan2018second,kuang2020voxel,yin2020lidar,deng2021voxel,mao2021voxel}, pillar-based~\cite{lang2018pointpillars,shi2022pillarnet,li2023pillarnext,zhou2023fastpillars} and raw point cloud-based~\cite{qi2018frustum,zhou2020joint,yang20203dssd,pan20213d}, etc. 
To evaluate the performance of existing detectors on our proposed cross-sensor benchmark, PointPillars, SECOND, PointRCNN, PV-RCNN and CenterPoint are utilized, and the details of the detectors are demonstrated in Sec. \ref{sec:detectors}.

\begin{figure*}[t!]
	\centering
	\includegraphics[width=0.73\textwidth]{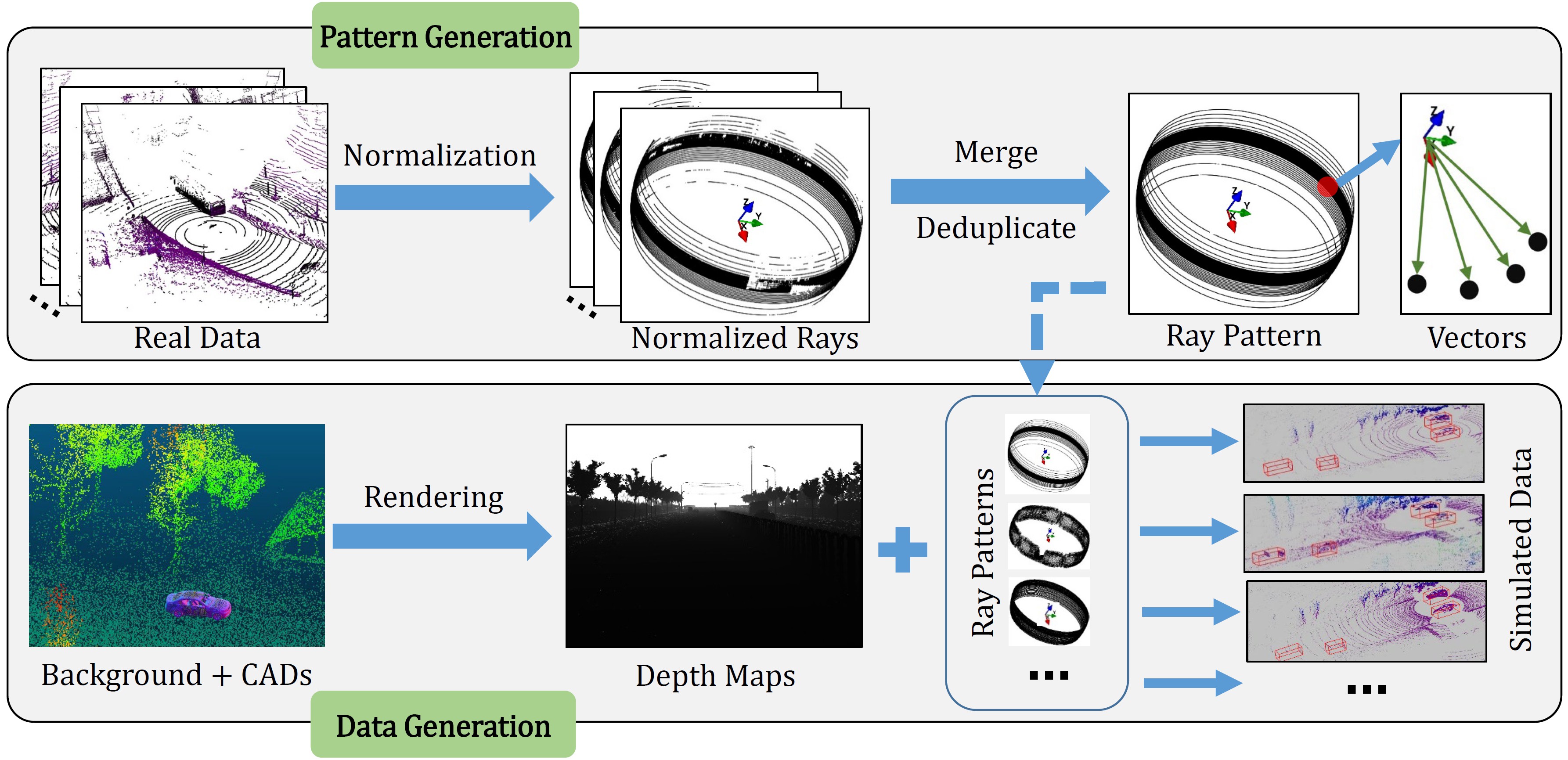} 
	\centering
	\caption{The overview of the proposed pattern-aware LiDAR Simulator framework. First of all, the real LiDAR points are normalized to a spherical surface. Due to points missing, statistics information from multiple scans is required to build the LiDAR \textit{Ray Pattern}. Then, the \textit{Ray Pattern} vectors are simultaneously projected and query the depth value from the depth map to generate the simulation point cloud.}
	\label{Fig:pipeline}
\end{figure*}

\section{LiDAR-CS Dataset}  \label{sec:method}

In this section, we introduce the LiDAR-CS Dataset along with the methodology for its construction. A pattern-aware LiDAR Simulator was designed to generate a realistic and high-quality dataset that can be used with most LiDAR sensors by inputting a few real scanned frames. 
Additionally, we provide information about the settings of the sensors, followed by the details of the dataset and evaluation metric.

\subsection{Cross-Sensor Data Generation}\label{subsec:cs_data_generation}
As previously mentioned, obtaining cross-sensor data with identical scenarios in real-world applications is a challenging task. To overcome this challenge, data simulation techniques can be employed to generate data. In order to reduce the domain gap between real and simulated data, we utilize an augmented LiDAR simulator proposed in \cite{Augmented_Sim_2020}. This simulator employs high-precision point cloud scans as a background and only synthesizes foreground objects with CAD models. Graphics rendering techniques are used to generate a realistic LiDAR point cloud. Our experimental results show that the generated point cloud performs similarly to real data for the 3D object detection task.

Furthermore, based on \cite{Augmented_Sim_2020}, we propose a pattern-aware LiDAR Simulation framework that is more general to suitable for different LiDAR sensors and combinations. The overview of the framework is shown in Fig. \ref{Fig:pipeline}, which consists of two modules, the \textit{Pattern Generation} module and the \textit{Data Generation} module.

\subsubsection{Pattern Generation Module}
For sensor simulation, parameters azimuth resolution, vertical angles, and spin rate are prerequisites. However, not all the LiDAR point clouds can find the corresponding parameters of the sensor, and multi-sensor equipment is common for nowadays AD applications, therefore it is complicated to search the parameters for all the sensors. Moreover, the parameters in the manual usually are physically different from the real applications.

To well handle this problem, we present a \textit{Pattern Generation} module, which can recover the LiDAR ray patterns from the scanned LiDAR points automatically and the LiDAR parameters are not prerequisites anymore. First of all, the point $p_i =  (x_i, y_i, z_i), i \leq N$,
(here $N$ denotes the number of points in this LiDAR frame) is normalized into a unit spherical surface by
\begin{equation}
    (\widetilde{x_i}, \widetilde{y_i}, \widetilde{z_i}) = \frac{(x_i, y_i, z_i)}{r},
\end{equation}
where 
\begin{equation}
r = \sqrt{{x_i}^2 + {y_i}^2 + {z_i}^2}.
\end{equation}
We call the normalized point cloud as \textit{Ray Pattern} which represents the distribution of the rays of this sensor. Note that some laser rays will not return (some points are missing) in a certain frame due to different reasons, therefore the \textit{Ray Pattern} from one frame may be incomplete. To handle this, we propose to merge the ray patterns across multi-frames and remove the duplicate rays to get a complete \textit{Ray Pattern}.

\subsubsection{Data Generation Module}

With the estimated \textit{Ray Pattern}, simulation LiDAR point cloud can be generated by leveraging the CG (Computer Graphics) techniques. 
We project the scene onto 6 faces of a cube centered at the LiDAR center to form the cube maps, the environment point cloud is rendered with surface splatting techniques to obtain the smooth and holeless projection while the obstacle CAD models are rendered with regular methods.

However, the rendering module is the speed bottleneck of the system during the data generation processing. For multi-sensors simulation, we can project all the \textit{Ray Patterns} to the depth map together and query the depth values simultaneously for each sensor. By this technique, the time-consuming point cloud simulation can remain nearly constant with the increasing number of sensors.
Since the location and attributes of the synthesis CAD models are predefined, the ground truth of the foreground objects is annotated automatically during the point cloud simulation processing. 

\begin{figure*}[t!]
	\centering
	\includegraphics[width=0.98\textwidth]{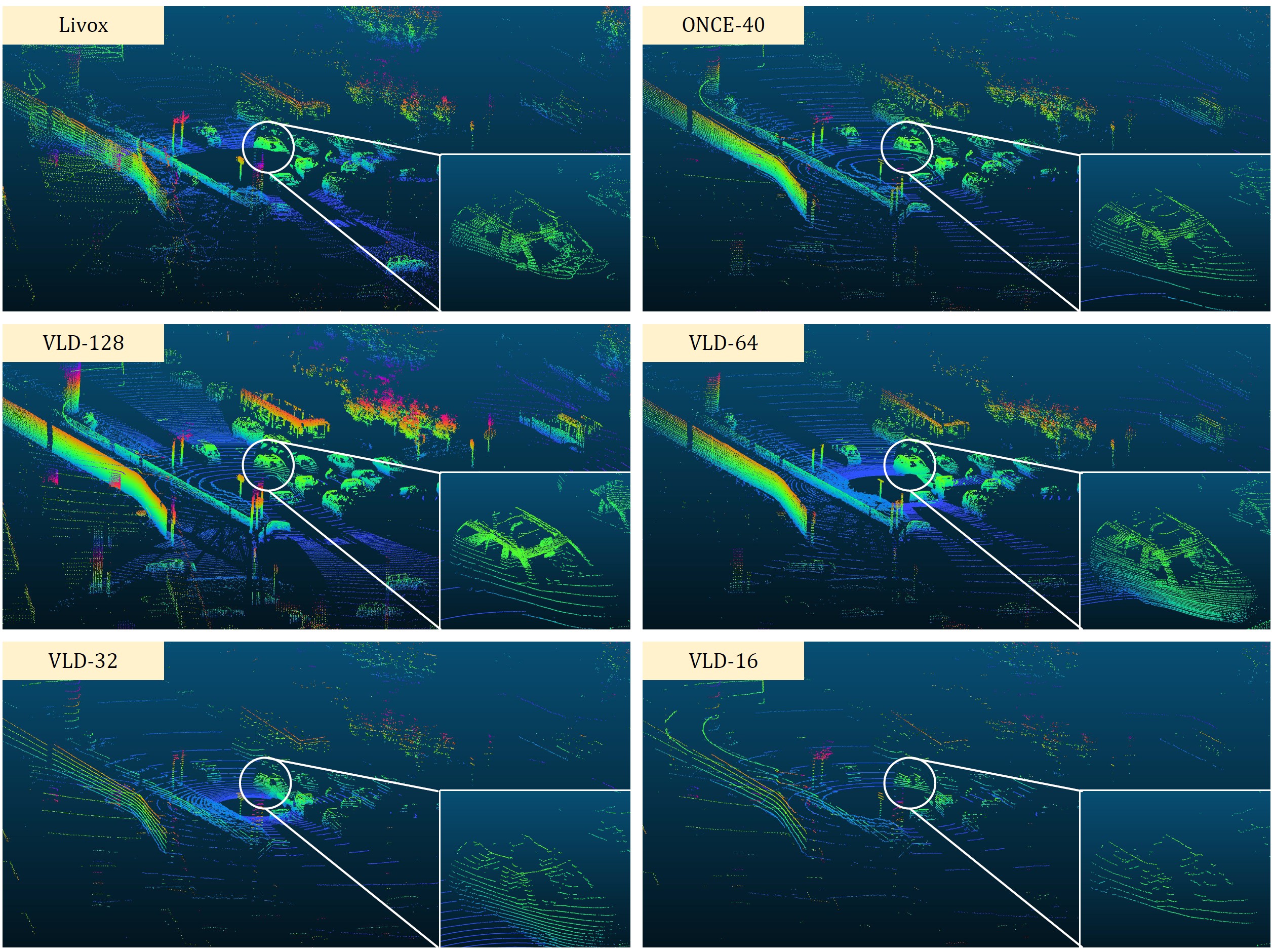} 
	\centering
	\caption{An example of the LiDAR-CS dataset. All the point clouds are generated from the same scenario under different sensor patterns. The points in the cycle are zoomed in and shown in the white boxes for a better view. The point clouds are colorized by the height of the points. Better viewed in color.}
	\label{Fig:vis}
\end{figure*}

\subsection{LiDAR-CS Dataset} \label{sec:device}

Currently, 6 different types of LiDAR sensors have been simulated here. All the sensors are set up at the same height (e.g., 2m in this work), and with $360^{\circ}$ horizontal field-of-view (FOV). The detailed information for each LiDAR sensor is listed below.    

\begin{enumerate}
    \item \textbf{Velodyne Puck 16} (VLD-16) has 16 channels and maximum 100m range with $30^{\circ}$ vertical FOV.
    
    \item \textbf{Velodyne 32E} (VLD-32) has 32 channels,  and vertical FOV from $-30^{\circ}$ to $+10^{\circ}$.
    
    \item \textbf{Velodyne HDL-64E} (VLD-64) has 64 channels, and 120m sensing range with $26.9^{\circ}$ vertical FOV. 
    
    \item \textbf{Velodyne 128} (VLD-128) can capture the range of maximum 245m, and vertical FOV from $-25^{\circ}$ to $+15^{\circ}$. 
    
    \item \textbf{ONCE-40} is used in  ONCE dataset \cite{mao2021one} which is a 40-beams LiDAR sensor with vertical FOV from $-25^{\circ}$ to $+15^{\circ}$. However, the parameters of the sensor have not been provided in the original paper \cite{mao2021one} which illustrates the necessity of our pattern-aware framework.

    \item \textbf{Livox.} Different from the mechanical scanning LiDAR devices introduced above, Livox is a solid-state LiDAR with incommensurable retina-like \cite{liu2021low} scanning patterns. We follow Livox Simu dataset \cite{link_livox_sim} and remove the dynamic pattern character to simplify the simulation processing. The sensor contains five Horizon LiDAR devices and one Tele-15 LiDAR device. 
    
\end{enumerate}

\subsection{Dataset Details}

We simulate 14,000 point cloud frames for each pattern from 6 different sensors. 
In total, LiDAR-CS Dataset contains 84,000 point cloud frames with 6 groups. We provide a semantic category and a cuboid bounding box as $(x, y, z, width, length, height, yaw~angle)$ for the foreground object annotation. For each group, the data is divided into two splits, 7,000 frames for training and 7,000 frames for testing. For better comparison, all the groups share the same data splitting. For the foreground objects, five categories have been annotated in this dataset as \textit{Car}, \textit{Truck}, \textit{Pedestrian}, \textit{Bicyclist} and \textit{Motorcyclist}. To be clear, the \textit{Truck} category includes not only trucks but also other big vehicles such as vans and buses, etc. The number of objects in each category has been provided in \tabref{tab:data_distribution}. Although the object number for ``Car'' is much larger than other categories, the object numbers for the rest categories are relatively balanced. 

\setlength{\tabcolsep}{4pt}
\begin{table}[h!]
\caption{Data distribution of annotated 3D objects in LiDAR-CS dataset.}
\label{tab:data_distribution}
\vspace{2mm}
\centering
\resizebox{0.46\textwidth}{!}{%
\begin{tabular}{c|ccccc}
\hline\noalign{\smallskip}
Types & Car & Truck & Pedestrian & Bicyclist & Motorcyclist \\
\noalign{\smallskip}
\hline\noalign{\smallskip}
3D Object & 1218k & 164k & 230k & 69k & 130k \\ \hline
\end{tabular}%
}
\end{table}
\setlength{\tabcolsep}{1.4pt}

In \figref{Fig:vis}, we illustrate an example of the LiDAR-CS data, in which all the background is from the same scenario and all the foreground objects are placed at the same location. In each sub-figure, we illustrate the LiDAR point cloud generated by different LiDAR patterns. For better visualization, a small part has been cropped and zoomed in to show the details of the object. 

\subsection{Evaluation Metric}

We follow the evaluation metric in ONCE benchmark \cite{mao2021one}. Mean average precision (mAP) is calculated over the five categories while the orientations of the objects are also considered, objects will be regarded as false positive when the orientation fails to fall in $-90\degree$ to $90\degree$ of the ground truth orientation. Only objects within 70m are evaluated in our metric.

\begin{table*}[]
\caption{Cross evaluation on LiDAR-CS benchmark under five different LiDAR sensor patterns with five baseline detectors. ``Ped.'' is short for ``Pedestrian''.}
\label{tab:cross_sensor}
\vspace{2mm}
\resizebox{\textwidth}{!}{%
\begin{tabular}{cc||cccc||cccc||cccc||cccc||cccc}
\hline
\multicolumn{2}{c||}{\diagbox{Train}{Val}} & \multicolumn{4}{c||}{Livox} & \multicolumn{4}{c||}{ONCE-40} & \multicolumn{4}{c||}{VLD-64} & \multicolumn{4}{c||}{VLD-32} & \multicolumn{4}{c}{VLD-16} \\ \hline \hline

\multicolumn{1}{c|}{} & Methods & \multicolumn{1}{c}{\textbf{mAP}} &  \multicolumn{1}{c}{\textbf{Car}} & \multicolumn{1}{c}{\textbf{Truck}} & \multicolumn{1}{c||}{\textbf{Ped.}} & \multicolumn{1}{c}{\textbf{mAP}} & \multicolumn{1}{c}{\textbf{Car}} & \multicolumn{1}{c}{\textbf{Truck}} & \multicolumn{1}{c||}{\textbf{Ped.}} & \multicolumn{1}{c}{\textbf{mAP}} &
\multicolumn{1}{c}{\textbf{Car}} & \multicolumn{1}{c}{\textbf{Truck}} & \multicolumn{1}{c||}{\textbf{Ped.}} & \multicolumn{1}{c}{\textbf{mAP}} &
\multicolumn{1}{c}{\textbf{Car}} & \multicolumn{1}{c}{\textbf{Truck}} & \multicolumn{1}{c||}{\textbf{Ped.}} & \multicolumn{1}{c}{\textbf{mAP}} & \multicolumn{1}{c}{\textbf{Car}} & \multicolumn{1}{c}{\textbf{Truck}} & \multicolumn{1}{c}{\textbf{Ped.}} \\ \cline{1-22} \hline \hline

\multicolumn{1}{c|}{\multirow{5}{*}{Livox}} & PointPillar  & 54.13 & 77.21 & 83.29 & 25.41 & 57.46 & 77.29 & 85.25 & 27.65 & 58.15 & 77.34 & 85.37 & 27.09 & 38.79 & 51.76 & 62.54 & 23.02 & 30.38 & 42.87 & 47.54 & 20.38 \\ \cline{2-22} 
\multicolumn{1}{c|}{} & SECOND & 57.39 & 75.56 & 83.15 & 25.54 & 61.04 & 77.04 & 85.28 & 27.25 & 61.52 & 76.97 & 85.20 & 26.67 & 40.14 & 49.48 & 60.63 & 22.89 & 29.39 & 38.88 & 42.89 & 18.51 \\ \cline{2-22} 
\multicolumn{1}{c|}{} & PointRCNN & 36.50 & 53.82 & 63.66 & 10.07 & 42.90 & 53.87 & 69.68 & 17.40 & 38.86 & 51.82 & 67.69 & 11.64 & 27.05 & 36.96 & 40.83 & 10.21 & 22.23 & 30.75 & 30.06 & 8.71 \\ \cline{2-22} 
\multicolumn{1}{c|}{} & PV-RCNN & 62.34 & 82.66 & 88.04 & 26.16 & 66.63 & 84.71 & 90.27 & 27.64 & 67.43 & 84.76 & 90.28 & 28.04 & 44.84 & 56.37 & 66.38 & 22.95 & 33.09 & 45.33 & 47.23 & 18.01 \\ \cline{2-22} 
\multicolumn{1}{c|}{} & CenterPoint & 68.55 & 79.31 & 83.41 & 49.36 & 73.75 & 81.05 & 85.55 & 54.37 & 74.66 & 81.23 & 85.82 & 55.15 & 48.20 & 51.10 & 60.20 & 39.47 & 35.54 & 40.45 & 43.23 & 30.57 \\ \hline \hline

\multicolumn{1}{c|}{\multirow{5}{*}{ONCE-40}} & PointPillar & 49.53 & 68.34 & 78.75 & 23.37 & 62.96 & 82.08 & 88.14 & 30.98 & 63.06 & 82.10 & 87.71 & 29.65 & 36.90 & 51.62 & 58.69 & 20.74 & 27.45 & 41.35 & 43.34 & 15.77 \\ \cline{2-22} 

\multicolumn{1}{c|}{} & SECOND & 53.30 & 66.76 & 76.72 & 26.92 & 66.14 & 81.80 & 87.83 & 31.61 & 66.46 & 81.88 & 87.93 & 31.13 & 35.22 & 45.34 & 53.67 & 20.35 & 23.45 & 33.76 & 34.16 & 13.41 \\ \cline{2-22} 
\multicolumn{1}{c|}{} & PointRCNN & 37.61 & 55.73 & 61.58 & 12.02 & 48.02 & 63.91 & 69.87 & 18.83 & 45.38 & 63.83 & 71.73 & 14.26 & 28.03 & 36.92 & 38.90 & 11.67 & 20.47 & 28.49 & 26.04 & 8.96 \\ \cline{2-22} 
\multicolumn{1}{c|}{} & PV-RCNN & 56.26 & 70.18 & 79.12 & 25.74 & 70.42 & 87.12 & 90.59 & 30.66 & 71.26 & 87.22 & 90.62 & 31.39 & 39.60 & 52.52 & 56.90 & 18.32 & 27.20 & 41.61 & 35.49 & 12.48 \\ \cline{2-22} 
\multicolumn{1}{c|}{} & CenterPoint & 61.92 & 68.61 & 76.73 & 46.33 & 77.19 & 84.23 & 88.36 & 58.17 & 77.50 & 84.20 & 88.09 & 58.97 & 42.07 & 48.77 & 53.98 & 32.77 & 29.26 & 37.22 & 35.81 & 22.86 \\ \hline \hline

\multicolumn{1}{c|}{\multirow{5}{*}{VLD-64}} & PointPillar & 50.50 & 70.42 & 78.69 & 23.95 & 62.77 & 82.03 & 87.74 & 30.54 & 64.16 & 82.19 & 88.04 & 31.76 & 36.29 & 51.50 & 57.98 & 19.45 & 26.42 & 40.96 & 41.85 & 14.82 \\ \cline{2-22} 
\multicolumn{1}{c|}{} & SECOND & 54.24 & 68.42 & 76.31 & 27.24 & 66.31 & 81.86 & 86.98 & 30.52 & 67.30 & 82.09 & 87.85 & 31.28 & 34.87 & 45.77 & 50.89 & 19.42 & 22.99 & 35.16 & 32.96 & 12.56 \\ \cline{2-22} 
\multicolumn{1}{c|}{} & PointRCNN & 36.01 & 51.77 & 57.61 & 10.22 & 44.72 & 59.92 & 63.84 & 16.35 & 41.69 & 57.89 & 63.80 & 13.69 & 27.56 & 35.26 & 38.80 & 12.51 & 21.72 & 28.87 & 27.86 & 9.70 \\ \cline{2-22} 
\multicolumn{1}{c|}{} & PV-RCNN & 57.39 & 72.15 & 81.03 & 25.32 & 70.55 & 88.91 & 90.77 & 29.68 & 71.97 & 89.15 & 90.95 & 30.74 & 39.94 & 54.38 & 57.49 & 17.51 & 27.47 & 42.16 & 37.60 & 11.29 \\ \cline{2-22} 
\multicolumn{1}{c|}{} & CenterPoint & 62.16 & 68.89 & 75.93 & 47.10 & 76.88 & 84.18 & 87.95 & 57.14 & 78.00 & 86.08 & 88.18 & 59.23 & 41.86 & 48.51 & 53.72 & 32.92 & 27.70 & 36.34 & 35.25 & 20.66 \\ \hline \hline

\multicolumn{1}{c|}{\multirow{5}{*}{VLD-32}} & PointPillar & 34.86 & 46.60 & 63.58 & 17.04 & 44.60 & 61.34 & 74.18 & 21.45 & 42.96 & 60.76 & 71.64 & 19.94 & 46.08 & 64.11 & 75.60 & 23.64 & 36.55 & 52.73 & 55.40 & 21.75 \\ \cline{2-22} 
\multicolumn{1}{c|}{} & SECOND & 34.42 & 40.22 & 55.37 & 20.33 & 45.77 & 60.21 & 69.42 & 22.87 & 41.28 & 56.69 & 63.00 & 20.76 & 48.40 & 62.01 & 73.77 & 26.45 & 37.89 & 50.44 & 54.04 & 22.08 \\ \cline{2-22} 
\multicolumn{1}{c|}{} & PointRCNN & 32.32 & 45.32 & 54.87 & 8.93 & 39.55 & 57.71 & 61.44 & 13.44 & 36.11 & 55.51 & 61.29 & 9.13 & 35.76 & 49.09 & 51.23 & 16.11 & 33.47 & 43.95 & 44.32 & 16.27 \\ \cline{2-22} 
\multicolumn{1}{c|}{} & PV-RCNN & 35.46 & 43.67 & 56.81 & 17.96 & 49.05 & 71.07 & 70.14 & 19.89 & 44.97 & 66.61 & 63.68 & 18.38 & 53.45 & 71.48 & 79.37 & 24.64 & 43.61 & 59.75 & 61.38 & 21.54 \\ \cline{2-22} 
\multicolumn{1}{c|}{} & CenterPoint & 41.84 & 42.97 & 55.43 & 25.73 & 55.14 & 62.47 & 69.69 & 36.68 & 51.23 & 59.74 & 67.01 & 34.52 & 56.94 & 63.55 & 73.26 & 41.50 & 43.83 & 51.56 & 52.35 & 34.28 \\ \hline \hline

\multicolumn{1}{c|}{\multirow{5}{*}{VLD-16}} & PointPillar & 22.52 & 29.71 & 42.82 & 12.62 & 28.41 & 40.27 & 49.54 & 14.80 & 25.50 & 35.13 & 46.32 & 13.84 & 35.83 & 53.70 & 54.45 & 19.27 & 39.70 & 57.09 & 64.36 & 22.64 \\ \cline{2-22} 
\multicolumn{1}{c|}{} & SECOND & 17.45 & 18.12 & 35.83 & 12.63 & 26.02 & 37.31 & 44.80 & 14.45 & 19.22 & 25.26 & 38.50 & 12.00 & 39.68 & 54.74 & 61.37 & 20.50 & 39.07 & 54.33 & 61.72 & 20.26 \\ \cline{2-22} 
\multicolumn{1}{c|}{} & PointRCNN & 25.63 & 34.35 & 49.05 & 4.76 & 36.83 & 51.33 & 60.79 & 10.06 & 33.38 & 48.78 & 62.37 & 5.55 & 38.47 & 49.09 & 56.73 & 17.92 & 37.45 & 46.43 & 49.93 & 19.61 \\ \cline{2-22} 
\multicolumn{1}{c|}{} & PV-RCNN & 16.21 & 18.19 & 37.62 & 8.15 & 26.05 & 41.23 & 50.46 & 10.22 & 17.65 & 26.66 & 40.64 & 5.12 & 46.18 & 65.20 & 72.13 & 21.27 & 46.28 & 64.43 & 69.61 & 22.63 \\ \cline{2-22} 
\multicolumn{1}{c|}{} & CenterPoint & 20.41 & 20.94 & 32.83 & 10.69 & 26.09 & 36.12 & 41.47 & 15.78 & 17.42 & 26.03 & 34.02 & 5.39 & 48.69 & 56.36 & 62.15 & 34.86 & 47.56 & 54.49 & 61.44 & 34.39 \\ \hline
\end{tabular}%
}

\end{table*}

\section{Empirical Studies} \label{sec:experiments}

On top of LiDAR-CS benchmark, we evaluate several 3D object detection baselines on the generalization ability and explore the influence of the points distribution domain gap. 

\subsection{3D Object Detection Baselines} \label{sec:detectors}

To well analyze the performances of different detectors across different sensor data, we employ five state-of-the-art 3D object detection baselines here which include points-based, voxel-based, and anchor-free-based frameworks respectively. Details of these baselines are introduced below: 
\begin{itemize}[leftmargin=*]
\item \textbf{PointPillars} \cite{lang2018pointpillars} transforms the points into vertical pillars to form the 2D feature map and the feature in each pillar is extracted with a light-weighted MLP network, then the 2D backbone is employed for the following detection task; 
\item \textbf{SECOND}~\cite{yan2018second} is an improved version of VoxelNet by introducing the sparse 3D convolution to replace the traditional 3D convolution which can boost both the 3D detection performance and efficiency; 
\item \textbf{PointRCNN}~\cite{shi2019pointrcnn} is a purely points-based framework that utilizes PointNet++ as the backbone to extract features for the segmentation branch and the bounding box proposal branches respectively. In addition, the second refine module is also designed to refine the bound box proposal and leverage the cropped point cloud together with the point features extracted from the first stage; 
\item \textbf{PV-RCNN}~\cite{shi2020pv} is a hybrid approach that learns discriminative features by combining both the 3D voxel and points representation; 
\item \textbf{CenterPoint}~\cite{yin2021center} is an anchor-free-based method and achieves state-of-the-art performance on various benchmarks.
\end{itemize}

\subsection{Cross-Sensors Evaluation}
Five mainstream detectors are employed here to reveal perception domain gaps between different LiDAR sensors. In our experiments, all the detectors are trained independently on the training split (2,000 frames of them are used as validation set during the training) of each point cloud group and evaluated on the testing split of all the groups, respectively. 

The detailed evaluation results are shown in \tabref{tab:cross_sensor}. Limited by the space, the results for VLD-128 sensor data and categories of ``Bicyclist'' and ``Motorcyclist'' have not been given in this table, readers could refer to our dataset homepage for the complete evaluation results. To be clear, the \textbf{mAP} is still computed over all the five categories.

\vspace{0.2cm}
\noindent\textbf{Domain Gaps from Points Distribution.} From this table, we can obviously find some interesting observations. \textbf{Firstly}, the model trained and tested on the same sensor gives the best performance in most situations. This is easy to understand because they come from the same sensor therefore the point cloud distribution is exactly identical. \textbf{Secondly}, the detectors trained on a low-resolution dataset can achieve comparable performance on the high-resolution testing dataset while the opposite is not true. Taking the detectors trained on VLD-32, they achieve similar performances by testing on VLD-32 and VLD-64 while the performances drop dramatically on VLD-16. \textbf{Thirdly}, the domain gaps positively correlate to the similarity of point ray distribution. For example, models trained on VLD-32 data outperform models trained on VLD-64 data when testing on VLD-16 dataset, and vice versa. \textbf{Finally}, we find that some sensors exhibit similar performance, such as ONCE-40 and VLD-64, and the mAP results are very close whether as training data or testing data. This is because even though they have different beam numbers, they share similar vertical angles' distribution.

\vspace{0.2cm}
\noindent\textbf{Detectors Evaluation.} We also compare the performances of different detectors. Overall, PV-RCNN achieves the best performance on large obstacles such as ``Car'' and ``Truck'', while CenterPoint outperforms other baseline detectors with a remarkable gap in the ``Pedestrian'' category. When facing the domain gap situation, the performances of all the baselines decrease dramatically, especially for PointRCNN which relies on a Points-based backbone for feature extraction and the extracted features will change dramatically with the points distribution change. Relatively speaking, voxel-based methods have the ability to deal with the change of point distribution to some extent. Experimental results illustrate that mainstream baseline detectors do not have the points-distribution-insensitive ability. We propose the hypothesis that the extracted features from the detectors are disabled to learn the real structure of the objects or underlying surfaces, but are confined to learning the distribution of points.

\subsection{Domain Alignment Evaluation}
To handle the domain gap between different sensor data, an intuitive way is to align the laser beams from the source sensor to the target sensor. To our knowledge, few approaches and datasets have been proposed to solve the cross-sensors gap problem. Two baseline domain alignment methods (e.g., ST3D \cite{yang2021st3d} and SN \cite{wang2020train}) are employed for evaluation here. SN takes the BBoxs' size from the target dataset to normalize the BBoxs in the source dataset during the training while ST3dD uses a \textit{Random Object Scaling} strategy to expand the diversity of object sizes. 

For comparison, we propose a simple strategy, NNDS (Nearest-neighbor scan down-sampling), which reduces the domain of point distribution by downsampling the source point cloud along the target scan distribution, as in \cite{alonso2020domain}. But different from \cite{alonso2020domain} simply dropping the scan lines uniformly, NNDS utilizes nearest-neighbor scan searching with target data as a reference and drops the other lines.

Tasks of VLD-64 $\rightarrow$ VLD-32 and VLD-64 $\rightarrow$ VLD-16 are given in Tab. \ref{tab:eval_st3d_1}. PointPillars and CenterPoint are employed here as baseline detectors. From the table, we can see that the baselines perform worse than the NNDS. Both \textit{ST3D} and \textit{SN} are proposed for solving the domain gap that comes from the different scenarios. But our dataset is proposed to simulate the domain gap coming from different LiDAR sensors and the statistical information of objects from different sensors are the same. Therefore, our dataset introduces a new challenge for domain alignment.

\begin{table}[h!]
\caption{Evaluation for cross-sensor domain alignment. Two detectors are used here: SECOND with IoU head and PV-RCNN. The improvement based on the source-only method is highlighted in blue.}
\label{tab:eval_st3d_1}
\vspace{2mm}
\centering\resizebox{0.46\textwidth}{!}{%
\begin{tabular}{l|c|cc}
\hline
\multicolumn{1}{c|}{\multirow{2}{*}{\bf{Task}}} & \multicolumn{1}{c|}{\multirow{2}{*}{\bf{Method}}} & \multicolumn{2}{c}{\bf{mAP} (\%)} \\ \cline{3-4} 
\multicolumn{1}{c|}{} & \multicolumn{1}{c|}{} & \multicolumn{1}{l|}{SECOND-IoU} & PV-RCNN \\ \hline \hline

\multirow{4}{*}{VLD-64 $\rightarrow$ VLD-32} & Source Only   & \multicolumn{1}{l|}{34.56}  &  \multicolumn{1}{l}{39.94}  \\ \cline{2-4} 
                                             & SN            & \multicolumn{1}{l|}{34.56 (\textbf{\B{0 $\uparrow$}})}  &  \multicolumn{1}{l}{39.94 (\textbf{\B{0 $\uparrow$}})}  \\ \cline{2-4} 
                                             
                                             & ST3D          & \multicolumn{1}{l|}{34.89 (\textbf{\B{0.33 $\uparrow$}})}    &     \multicolumn{1}{l}{40.43 (\textbf{\B{0.49 $\uparrow$}})}    \\ \cline{2-4} 
                                             
                                             & NNDS          & \multicolumn{1}{l|}{\bf{48.02} (\textbf{\B{13.46 $\uparrow$}})}  &  \multicolumn{1}{l}{\bf{52.29} (\textbf{\B{12.35 $\uparrow$}})} \\ 
                                            \hline \hline
                                             
\multirow{4}{*}{VLD-64 $\rightarrow$ VLD-16} & Source Only   & \multicolumn{1}{l|}{23.01}  &  \multicolumn{1}{l}{27.47}  \\ \cline{2-4} 
                                             & SN            & \multicolumn{1}{l|}{23.01  (\textbf{\B{0 $\uparrow$}})}  &  \multicolumn{1}{l}{27.47  (\textbf{\B{0 $\uparrow$}})}  \\ \cline{2-4} 
                                             
                                             & ST3D          & \multicolumn{1}{l|}{22.95 (\textbf{\B{-0.06 $\uparrow$}})}       &  \multicolumn{1}{l}{27.78 (\textbf{\B{0.31 $\uparrow$}})}        \\ \cline{2-4} 
                                             
                                            & NNDS          & \multicolumn{1}{l|}{\bf{39.76} (\textbf{\B{16.75 $\uparrow$}})}  &  \multicolumn{1}{l}{\bf{44.85} (\textbf{\B{17.38 $\uparrow$}})}    \\ 
                                            \hline
\end{tabular}%
}
\end{table}

\section{Potential Applications and Discussion} \label{sec:discuss}

The potential applications of the LiDAR-CS dataset are virtually limitless. Although originally designed for 3D object detection, it could also be repurposed for other purposes. 

\vspace{4pt}

\noindent\textbf{Data Domain Adaptation} The domain gaps between different sensors, simulation, and real-world, are very important topics for real AD applications. The proposed pattern-aware point cloud simulation technique can be employed for generating any type of LiDAR sensor. 

\noindent\textbf{Distribution-insensitive features.} Our LiDAR-CS dataset could be applied to evaluate the anti-interference ability of features extracted from different backbones, as we showed in Tab. \ref{tab:cross_sensor}. Furthermore, perception systems with sensor-invariant ability can also be evaluated.

\noindent\textbf{Large-scale Point Cloud Up-sampling or Re-sampling.} While there are many methods for up-sampling point clouds, most have only been tested on small samples from CAD models with a few thousand points. However, few studies have focused on up-sampling LiDAR point clouds in large-scale autonomous driving scenarios. Our dataset includes realistic point clouds from various scenarios and large-scale frames with different resolutions, making it an ideal resource for researching point cloud up-sampling and re-sampling.

\noindent\textbf{Semantic Segmentation.} Since the ground truth bounding box of all objects is provided, obtaining the semantic label for foreground objects is easy. Our dataset can also be used for semantic segmentation.

\noindent\textbf{Sensor Selection.} Our approach can also be used to select LiDAR hardware and adjust parameters. Instead of relying on an intuitive selection of device type and installation parameters, or choosing device combinations, and then collecting and annotating data, and training perception models, it would be more scientific and safer to evaluate devices and settings with perception systems beforehand using low-cost simulation data.

\section{Conclusion}

In this paper, we propose LiDAR-CS Dataset, which fills in the missing benchmark on points distribution domain adaptation and sensor-insensitive detectors for 3D object detection. For generating LiDAR-CS Dataset, we present a \textit{pattern-aware LiDAR Simulator} to simplify the calculation of sensor ray tracing and accelerate the data generation. Through empirical experiments on baseline detectors using various sensor devices, we have gained insightful conclusions and heuristic understanding which can positively impact future research. Our dataset has been made available to the public, with the hope that it will encourage further research on 3D point cloud technology. 

\section{Acknowledgements}
This work was supported in part by GuangDong Basic and Applied Basic Research Foundation No. 2020B1515130004 and FDCT Grants 0123/2022/AFJ, 0081/2022/A2.

{\small
\bibliographystyle{IEEEtran}
\bibliography{ref}
}

\end{document}